\documentclass[10pt,twocolumn,letterpaper]{article}

\usepackage{cvpr}
\usepackage{times}
\usepackage{epsfig}
\usepackage{wrapfig}
\usepackage{bm}
\usepackage[utf8]{inputenc} 
\usepackage[T1]{fontenc}    
\usepackage{url}            
\usepackage{booktabs}       
\usepackage{amsfonts}       
\usepackage{nicefrac}       
\usepackage{microtype}      
\usepackage{graphicx}
\usepackage{amsmath,amssymb}
\usepackage{amsthm}
\usepackage{caption}
\usepackage{color}
\usepackage{url}
\newtheorem{property}{Property}
\newtheorem{definition}{Definition}

\usepackage[pagebackref=true,breaklinks=true,letterpaper=true,colorlinks,bookmarks=false]{hyperref}
\cvprfinalcopy 

\def\cvprPaperID{76} 

\ifcvprfinal\fi
\begin{document}

\title{SphereFace: Deep Hypersphere Embedding for Face Recognition}

\author{Weiyang Liu\textsuperscript{1}\ \ \ \ \ Yandong Wen\textsuperscript{2}\ \ \ \ \ Zhiding Yu\textsuperscript{2}\ \ \ \ \ Ming Li\textsuperscript{3}\ \ \ \ \ Bhiksha Raj\textsuperscript{2}\ \ \ \ \ Le Song\textsuperscript{1}\\
  \textsuperscript{1}Georgia Institute of Technology \ \ \ \ \ \textsuperscript{2}Carnegie Mellon University \ \ \ \ \ \textsuperscript{3}Sun Yat-Sen University\\
{\tt\small wyliu@gatech.edu, \{yandongw,yzhiding\}@andrew.cmu.edu, lsong@cc.gatech.edu}
}
\maketitle
\thispagestyle{empty}
\begin{abstract}

This paper addresses deep face recognition (FR) problem under open-set protocol, where ideal face features are expected to have smaller maximal intra-class distance than minimal inter-class distance under a suitably chosen metric space. However, few existing algorithms can effectively achieve this criterion. To this end, we propose the angular softmax (A-Softmax) loss that enables convolutional neural networks (CNNs) to learn angularly discriminative features. Geometrically, A-Softmax loss can be viewed as imposing discriminative constraints on a hypersphere manifold, which intrinsically matches the prior that faces also lie on a manifold. Moreover, the size of angular margin can be quantitatively adjusted by a parameter $m$. We further derive specific $m$ to approximate the ideal feature criterion. Extensive analysis and experiments on Labeled Face in the Wild (LFW), Youtube Faces (YTF) and MegaFace Challenge show the superiority of A-Softmax loss in FR tasks. The code has also been made publicly available\footnote{See the code at  \url{https://github.com/wy1iu/sphereface}.}.
\end{abstract}
\vspace{-0.5mm}
\section{Introduction}
\begin{figure}[t]
  \centering
  \renewcommand{\captionlabelfont}{\footnotesize}
  \setlength{\abovecaptionskip}{3pt}
  \setlength{\belowcaptionskip}{-12pt}
  \includegraphics[width=3.05in]{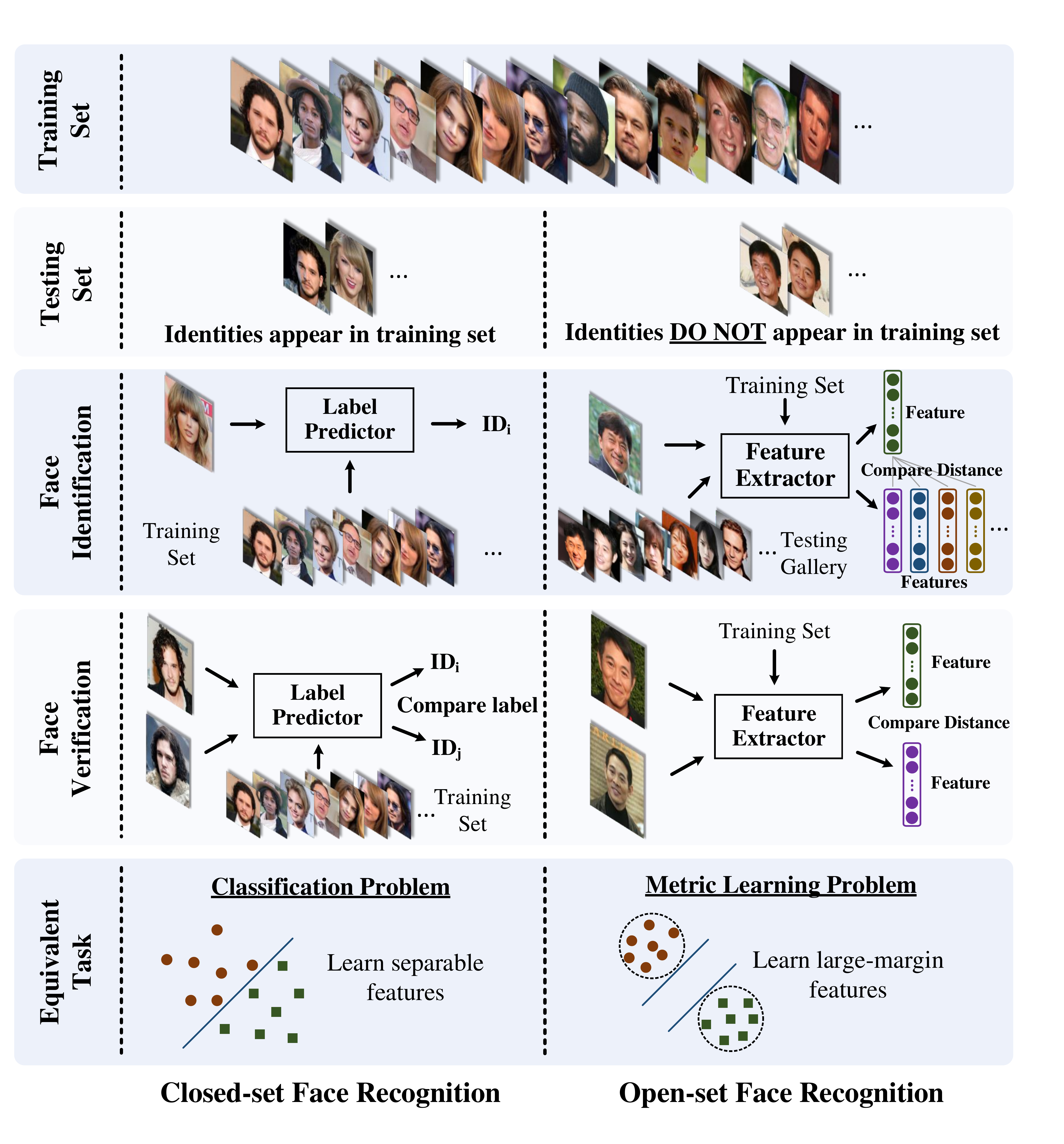}\\
  \caption{\footnotesize Comparison of open-set and closed-set face recognition.}\label{fr}
\end{figure}
Recent years have witnessed the great success of convolutional neural networks (CNNs) in face recognition (FR). Owing to advanced network architectures \cite{krizhevsky2012imagenet,simonyan2014very,szegedy2015going,he2016deep} and discriminative learning approaches \cite{sun2014deep2,schroff2015facenet,wen2016discriminative}, deep CNNs have boosted the FR performance to an unprecedent level. Typically, face recognition can be categorized as face identification and face verification \cite{huang2014labeled,Kemelmacher-Shlizerman_2016_CVPR}. The former classifies a face to a specific identity, while the latter determines whether a pair of faces belongs to the same identity.
\par
In terms of testing protocol, face recognition can be evaluated under closed-set or open-set settings, as illustrated in Fig.~\ref{fr}. For closed-set protocol, all testing identities are predefined in training set. It is natural to classify testing face images to the given identities. In this scenario, face verification is equivalent to performing identification for a pair of faces respectively (see left side of Fig.~\ref{fr}). Therefore, closed-set FR can be well addressed as a classification problem, where features are expected to be separable. For open-set protocol, the testing identities are usually disjoint from the training set, which makes FR more challenging yet close to practice. Since it is impossible to classify faces to known identities in training set, we need to map faces to a discriminative feature space. In this scenario, face identification can be viewed as performing face verification between the probe face and every identity in the gallery (see right side of Fig.~\ref{fr}). Open-set FR is essentially a metric learning problem, where the key is to learn discriminative large-margin features.
\par

Desired features for open-set FR are expected to satisfy the
 criterion that the maximal intra-class distance is smaller than the minimal inter-class distance under a certain metric space. This criterion is necessary if we want to achieve perfect accuracy using nearest neighbor. However, learning features with this criterion is generally difficult because of the intrinsically large intra-class variation and high inter-class similarity \cite{ross2004multimodal} that faces exhibit.
\par
\begin{figure*}[t]
  \centering
  \renewcommand{\captionlabelfont}{\footnotesize}
  \setlength{\abovecaptionskip}{4pt}
  \setlength{\belowcaptionskip}{-10pt}
  \includegraphics[width=6.85in]{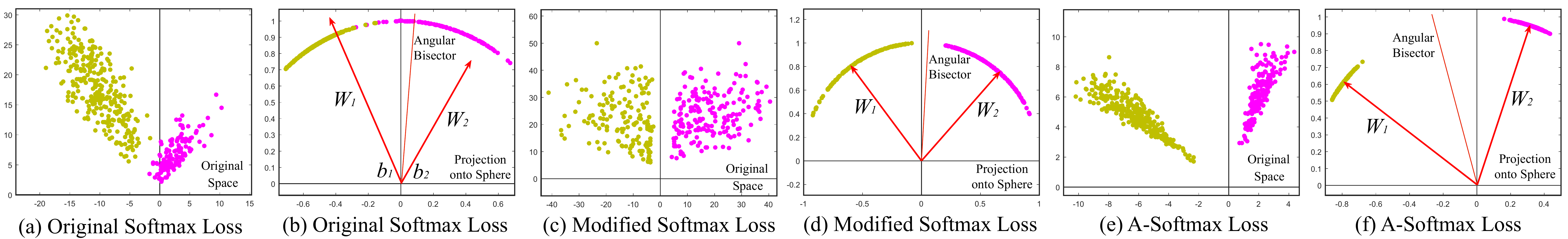}\\
  \caption{\footnotesize Comparison among softmax loss, modified softmax loss and A-Softmax loss. In this toy experiment, we construct a CNN to learn 2-D features on a subset of the CASIA face dataset. In specific, we set the output dimension of FC1 layer as 2 and visualize the learned features. Yellow dots represent the first class face features, while purple dots represent the second class face features. One can see that features learned by the original softmax loss can not be classified simply via angles, while modified softmax loss can. Our A-Softmax loss can further increase the angular margin of learned features.}\label{comp}
\end{figure*}
\par
Few CNN-based approaches are able to effectively formulate the aforementioned criterion in loss functions. Pioneering work \cite{taigman2014deepface,sun2014deep} learn face features via the softmax loss\footnote{Following \cite{liu2016large}, we define the softmax loss as the combination of the last fully connected layer, softmax function and cross-entropy loss.}, but softmax loss only learns separable features that are not discriminative enough. To address this, some methods combine softmax loss with contrastive loss \cite{sun2014deep2,sun2016sparsifying} or center loss \cite{wen2016discriminative} to enhance the discrimination power of features. \cite{schroff2015facenet} adopts triplet loss to
supervise the embedding learning, leading to state-of-the-art face recognition results. However, center loss only explicitly encourages intra-class compactness. Both contrastive loss \cite{hadsell2006dimensionality} and triplet loss \cite{schroff2015facenet} can not constrain on each individual sample, and thus require carefully designed pair/triplet mining procedure, which is both time-consuming and performance-sensitive.

\par
It seems to be a widely recognized choice to impose Euclidean margin to learned features, but a question arises: \emph{Is Euclidean margin always suitable for learning discriminative face features?}
To answer this question, we first look into how Euclidean margin based losses are applied to FR.

Most recent approaches \cite{sun2014deep2,sun2016sparsifying,wen2016discriminative} combine Euclidean margin based losses with softmax loss to construct a joint supervision. However, as can be observed from Fig.~\ref{comp}, the features learned by softmax loss have intrinsic angular distribution (also verified by \cite{wen2016discriminative}). In some sense, Euclidean margin based losses are incompatible with softmax loss, so it is not well motivated to combine these two type of losses.
\par
In this paper, we propose to incorporate angular margin instead. We start with a binary-class case to analyze the softmax loss. The decision boundary in softmax loss is $\thickmuskip=2mu \medmuskip=2mu (\bm{W}_{1} - \bm{W}_{2})\bm{x} + b_{1} - b_{2} = 0$, where $\bm{W}_{i}$ and $b_{i}$ are weights and bias\footnote{If not specified, the weights and biases in the paper are corresponding to the fully connected layer in the softmax loss.} in softmax loss, respectively. If we define $\bm{x}$ as a feature vector and constrain $\thickmuskip=2mu \medmuskip=2mu \|\bm{W}_1\| = \|\bm{W}_2\| = 1$ and $\thickmuskip=2mu \medmuskip=2mu b_{1}=b_{2} = 0$, the decision boundary becomes $\thickmuskip=2mu \medmuskip=2mu \|\bm{x}\|(\cos (\theta_1) -\cos (\theta_2)) = 0$, where $\theta_i$ is the angle between $\bm{W}_{i}$ and $\bm{x}$. The new decision boundary only depends on $\theta_1$ and $\theta_2$. Modified softmax loss is able to directly optimize angles, enabling CNNs to learn angularly distributed features (Fig.~\ref{comp}).

\par
Compared to original softmax loss, the features learned by modified softmax loss are angularly distributed, but not necessarily more discriminative. To the end, we generalize the modified softmax loss to angular softmax (A-Softmax) loss. Specifically, we introduce an integer $m$ ($\thickmuskip=2mu \medmuskip=2mu m \geq 1$) to quantitatively control the decision boundary. In binary-class case, the decision boundaries for class 1 and class 2 become $\thickmuskip=0mu \medmuskip=0mu \|\bm{x}\|(\cos (m\theta_1) -\cos (\theta_2)) = 0$ and $\thickmuskip=0mu \medmuskip=0mu \|\bm{x}\|(\cos (\theta_1) -\cos (m\theta_2)) = 0$, respectively. $m$ quantitatively controls the size of angular margin. Furthermore, A-Softmax loss can be easily generalized to multiple classes, similar to softmax loss. By optimizing A-Softmax loss, the decision regions become more separated, simultaneously enlarging the inter-class margin and compressing the intra-class angular distribution.
\par
A-Softmax loss has clear geometric interpretation. Supervised by A-Softmax loss, the learned features construct a discriminative angular distance metric that is equivalent to geodesic distance on a hypersphere manifold. A-Softmax loss can be interpreted as constraining learned features to be discriminative on a hypersphere manifold, which intrinsically matches the prior that face images lie on a manifold \cite{lee2003video,he2005face,talwalkar2008large}. The close connection between A-Softmax loss and hypersphere manifolds makes the learned features more effective for face recognition. For this reason, we term the learned features as \emph{SphereFace}.

\par
Moreover, A-Softmax loss can quantitatively adjust the angular margin via a parameter $m$, enabling us to do quantitative analysis. In the light of this, we derive lower bounds for the parameter $m$ to approximate the desired open-set FR criterion that the maximal intra-class distance should be smaller than the minimal inter-class distance.
\par
Our major contributions can be summarized as follows:
\par
(1) We propose A-Softmax loss for CNNs to learn discriminative face features with clear and novel geometric interpretation. The learned features discriminatively span on a hypersphere manifold, which intrinsically matches the prior that faces also lie on a manifold.
\par
(2) We derive lower bounds for $m$ such that A-Softmax loss can approximate the learning task that minimal inter-class distance is larger than maximal intra-class distance.
\par
(3) We are the very first to show the effectiveness of angular margin in FR. Trained on publicly available CASIA dataset \cite{yi2014learning}, \emph{SphereFace} achieves competitive results on several benchmarks, including Labeled Face in the Wild (LFW), Youtube Faces (YTF) and MegaFace Challenge 1.
\par
\section{Related Work}
\textbf{Metric learning.} Metric learning aims to learn a similarity (distance) function. Traditional metric learning \cite{xing2003distance,weinberger2005distance,kostinger2012large,ying2012distance} usually learns a matrix $\bm{A}$ for a distance metric $\thickmuskip=2mu \medmuskip=2mu \|\bm{x}_1-\bm{x}_2\|_{\bm{A}}=\sqrt{(\bm{x}_1-\bm{x}_2)^T\bm{A}(\bm{x}_1-\bm{x}_2)}$ upon the given features $\bm{x}_1,\bm{x}_2$. Recently, prevailing deep metric learning \cite{hu2014discriminative,lu2015multi,song2016deep,taigman2014deepface,sun2014deep2,schroff2015facenet,wen2016discriminative} usually uses neural networks to automatically learn discriminative features $\thickmuskip=2mu \medmuskip=2mu \bm{x}_1,\bm{x}_2$ followed by a simple distance metric such as Euclidean distance $\thickmuskip=2mu \medmuskip=2mu \|\bm{x}_1-\bm{x}_2\|_2$. Most widely used loss functions for deep metric learning are contrastive loss \cite{chopra2005learning,hadsell2006dimensionality} and triplet loss \cite{wang2014learning,schroff2015facenet,hoffer2014deep}, and both impose Euclidean margin to features.
\par
\textbf{Deep face recognition.} Deep face recognition is arguably one of the most active research area in the past few years. \cite{taigman2014deepface,sun2014deep} address the open-set FR using CNNs supervised by softmax loss, which essentially treats open-set FR as a multi-class classification problem. \cite{sun2014deep2} combines contrastive loss and softmax loss to jointly supervise the CNN training, greatly boosting the performance. \cite{schroff2015facenet} uses triplet loss to learn a unified face embedding. Training on nearly 200 million face images, they achieve current state-of-the-art FR accuracy. Inspired by linear discriminant analysis, \cite{wen2016discriminative} proposes center loss for CNNs and also obtains promising performance. In general, current well-performing CNNs \cite{sun2016sparsifying,liu2015targeting} for FR are mostly built on either contrastive loss or triplet loss. One could notice that state-of-the-art FR methods usually adopt ideas (e.g. contrastive loss, triplet loss) from metric learning, showing open-set FR could be well addressed by discriminative metric learning.
\par
L-Softmax loss \cite{liu2016large} also implicitly involves the concept of angles. As a regularization method, it shows great improvement on closed-set classification problems. Differently, A-Softmax loss is developed to learn discriminative face embedding. The explicit connections to hypersphere manifold makes our learned features particularly suitable for open-set FR problem, as verified by our experiments. In addition, the angular margin in A-Softmax loss is explicitly imposed and  can be quantitatively controlled (e.g. lower bounds to approximate desired feature criterion), while \cite{liu2016large} can only be analyzed qualitatively.

\section{Deep Hypersphere Embedding}
\subsection{Revisiting the Softmax Loss}
We revisit the softmax loss by looking into the decision criteria of softmax loss. In binary-class case, the posterior probabilities obtained by softmax loss are
\begin{equation}
\footnotesize
\thickmuskip=1mu p_{1} = \frac{\exp({\bm{W}_{1}^T\bm{x}+b_{1}})}{\exp({\bm{W}_{1}^T\bm{x}+b_{1}}) + \exp({\bm{W}^T_{2}\bm{x}+b_{2}})}
\end{equation}
\begin{equation}
\footnotesize
\thickmuskip=1mu p_{2} = \frac{\exp({\bm{W}^T_{2}\bm{x}+b_{2}})}{\exp({\bm{W}_{1}^T\bm{x}+b_{1}}) + \exp({\bm{W}_{2}^T\bm{x}+b_{2}})}
\end{equation}
 where $\bm{x}$ is the learned feature vector. $\bm{W}_{i}$ and $b_{i}$ are weights and bias of last fully connected layer  corresponding to class $i$, respectively. The predicted label will be assigned to class 1 if $\thickmuskip=2mu \medmuskip=2mu p_{1}>p_{2}$ and class 2 if $\thickmuskip=2mu \medmuskip=2mu p_{1}<p_{2}$. By comparing $p_{1}$ and $p_{2}$, it is clear that $\thickmuskip=2mu \medmuskip=2mu \bm{W}^T_{1}\bm{x}+b_{1}$ and $\thickmuskip=2mu \medmuskip=2mu \bm{W}^T_{2}\bm{x}+b_{2}$ determine the classification result. The decision boundary is $\thickmuskip=2mu \medmuskip=2mu (\bm{W}_{1} - \bm{W}_{2})\bm{x} + b_{1} - b_{2} = 0$. We then rewrite $\thickmuskip=2mu \medmuskip=2mu \bm{W}^T_{i}\bm{x}+b_{i}$ as $\thickmuskip=2mu \medmuskip=2mu \|\bm{W}^T_{i}\|\|\bm{x}\|\cos(\theta_i)+b_{i}$ where $\theta_{i}$ is the angle between $\bm{W}_{i}$ and $\bm{x}$. Notice that if we normalize the weights and zero the biases ($\thickmuskip=2mu \medmuskip=2mu \|\bm{W}_{i}\|=1$, $\thickmuskip=1mu b_{i}=0$), the posterior probabilities become $\thickmuskip=0mu \medmuskip=0mu p_{1} =\|\bm{x}\|\cos(\theta_{1})$ and $\thickmuskip=0mu \medmuskip=0mu p_{2} =\|\bm{x}\|\cos(\theta_{2})$. Note that $p_{1}$ and $p_{2}$ share the same $\bm{x}$, the final result only depends on the angles $\theta_{1}$ and $\theta_{2}$. The decision boundary also becomes $\thickmuskip=0mu \medmuskip=0mu \cos(\theta_1)-\cos(\theta_2)=0$ (i.e. angular bisector of vector $\bm{W}_1$ and $\bm{W}_2$). Although the above analysis is built on binary-calss case, it is trivial to generalize the analysis to multi-class case. During training, the modified softmax loss ($\thickmuskip=1mu \|\bm{W}_{i}\|=1,b_{i}=0$) encourages features from the $i$-th class to have smaller angle $\theta_{i}$ (larger cosine distance) than others, which makes angles between $\bm{W}_{i}$ and features a reliable metric for classification.
\par
To give a formal expression for the modified softmax loss, we first define the input feature $\bm{x}_i$ and its label $y_i$. The original softmax loss can be written as
\begin{equation}\label{sm1}
\footnotesize
L=\frac{1}{N}\sum_iL_i=\frac{1}{N}\sum_i-\log\big( \frac{e^{f_{y_i}}}{\sum_je^{f_j}} \big)
\end{equation}
where $f_j$ denotes the $j$-th element ($j\in[1,K]$, $K$ is the class number) of the class score vector $\bm{f}$, and $N$ is the number of training samples. In CNNs, $\bm{f}$ is usually the output of a fully connected layer $\bm{W}$, so $f_j=\bm{W}_j^T\bm{x}_i+b_j$ and $f_{y_i}=\bm{W}_{y_i}^T\bm{x}_i+b_{y_i}$ where $\bm{x}_i$, $\bm{W}_{j}, \bm{W}_{y_i}$ are the $i$-th training sample, the $j$-th and $y_i$-th column of $\bm{W}$ respectively. We further reformulate $L_i$ in Eq. \eqref{sm1} as
\begin{equation}\label{sm2}
\footnotesize
\begin{aligned}
L_i=&-\log\big( \frac{e^{\bm{W}_{y_i}^T\bm{x}_i+b_{y_i}}}{\sum_je^{\bm{W}_j^T\bm{x}_i+b_j}} \big)\\
=&-\log\big( \frac{e^{\|\bm{W}_{y_i}\|\|\bm{x}_i\|\cos(\theta_{y_i,i})+b_{y_i}}}{\sum_je^{\|\bm{W}_j\|\|\bm{x}_i\|\cos(\theta_{j,i})+b_j}} \big)
\end{aligned}
\end{equation}
in which $\thickmuskip=2mu \medmuskip=2mu \theta_{j,i}(0\leq\theta_{j,i}\leq\pi)$ is the angle between vector $\bm{W}_j$ and $\bm{x}_i$. As analyzed above, we first normalize $\thickmuskip=2mu \medmuskip=2mu \|\bm{W}_j\|=1,\forall j$ in each iteration and zero the biases. Then we have the modified softmax loss:
\begin{equation}\label{modified}
\footnotesize
L_{\textnormal{modified}}=\frac{1}{N}\sum_i-\log\big( \frac{e^{\|\bm{x}_i\|\cos(\theta_{y_i,i})}}{
\sum_{j}e^{\|\bm{x}_i\|\cos(\theta_{j,i})}} \big)
\end{equation}
Although we can learn features with angular boundary with the modified softmax loss, these features are still not necessarily discriminative. Since we use angles as the distance metric, it is natural to incorporate angular margin to learned features in order to enhance the discrimination power. To this end, we propose a novel way to combine angular margin.
\subsection{Introducing Angular Margin to Softmax Loss}
Instead of designing a new type of loss function and constructing a weighted combination with softmax loss (similar to contrastive loss) , we propose a more natural way to learn angular margin. From the previous analysis of softmax loss, we learn that decision boundaries can greatly affect the feature distribution, so our basic idea is to manipulate decision boundaries to produce angular margin. We first give a motivating binary-class example to explain how our idea works.
\par
Assume a learned feature $\bm{x}$ from class 1 is given and $\theta_i$ is the angle between $\bm{x}$ and $\bm{W}_i$, it is known that the modified softmax loss requires $\thickmuskip=2mu \medmuskip=2mu \cos(\theta_1)>\cos(\theta_2)$ to correctly classify $\bm{x}$. But what if we instead require $\thickmuskip=2mu \medmuskip=2mu \cos(m\theta_1)>\cos(\theta_2)$ where $\thickmuskip=2mu m\geq2$ is a integer in order to correctly classify $\bm{x}$? It is essentially making the decision more stringent than previous, because we require a lower bound\footnote{The inequality $\thickmuskip=2mu \medmuskip=2mu \cos(\theta_1)>\cos(m\theta_1)$ holds while $\thickmuskip=2mu \theta_1\in[0,\frac{\pi}{m}], m\geq2$.} of $\cos(\theta_1)$ to be larger than $\cos(\theta_2)$. The decision boundary for class 1 is $\thickmuskip=2mu \medmuskip=2mu \cos(m\theta_1)=\cos(\theta_2)$. Similarly, if we require $\thickmuskip=2mu \medmuskip=2mu \cos(m\theta_2)>\cos(\theta_1)$ to correctly classify features from class 2, the decision boundary for class 2 is $\thickmuskip=2mu \medmuskip=2mu \cos(m\theta_2)=\cos(\theta_1)$. Suppose all training samples are correctly classified, such decision boundaries will produce an angular margin of $\frac{m-1}{m+1}\theta_{2}^{1}$ where $\theta_{2}^{1}$ is the angle between $\bm{W}_1$ and $\bm{W}_2$. From angular perspective, correctly classifying $\bm{x}$ from identity 1 requires $\thickmuskip=2mu \theta_1<\frac{\theta_2}{m}$, while correctly classifying $\bm{x}$ from identity 2 requires $\thickmuskip=2mu \theta_2<\frac{\theta_1}{m}$. Both are more difficult than original $\thickmuskip=2mu \theta_1<\theta_2$ and $\thickmuskip=2mu \theta_2<\theta_1$, respectively. By directly formulating this idea into the modified softmax loss Eq. \eqref{modified}, we have
\begin{equation}\label{naiveangular}
\footnotesize
L_{\textnormal{ang}}=\frac{1}{N}\sum_i-\log\big( \frac{e^{\|\bm{x}_i\|\cos(m\theta_{y_i,i})}}{e^{\|\bm{x}_i\|\cos(m\theta_{y_i,i})}+
\sum_{j\neq y_i}e^{\|\bm{x}_i\|\cos(\theta_{j,i})}} \big)
\end{equation}
where $\theta_{y_i,i}$ has to be in the range of $[0,\frac{\pi}{m}]$. In order to get rid of this restriction and make it optimizable in CNNs, we expand the definition range of $\cos(\theta_{y_i,i})$ by generalizing it to a monotonically decreasing angle function $\psi(\theta_{y_i,i})$ which should be equal to $\cos(\theta_{y_i,i})$ in $[0,\frac{\pi}{m}]$. Therefore, our proposed A-Softmax loss is formulated as:
\begin{equation}\label{angular}
\footnotesize
L_{\textnormal{ang}}=\frac{1}{N}\sum_i-\log\big( \frac{e^{\|\bm{x}_i\|\psi(\theta_{y_i,i})}}{e^{\|\bm{x}_i\|\psi(\theta_{y_i,i})}+
\sum_{j\neq y_i}e^{\|\bm{x}_i\|\cos(\theta_{j,i})}} \big)
\end{equation}
in which we define $\thickmuskip=2mu \medmuskip=2mu \psi(\theta_{y_i,i})=(-1)^k\cos(m\theta_{y_i,i})-2k$, $ \theta_{y_i,i}\in[\frac{k\pi}{m},\frac{(k+1)\pi}{m}]$ and $\thickmuskip=2mu k\in[0,m-1]$. $\thickmuskip=2mu m\geq1$ is an integer that controls the size of angular margin. When $\thickmuskip=2mu m=1$, it becomes the modified softmax loss.
\par
\begin{table}[t]
\centering
\renewcommand{\captionlabelfont}{\footnotesize}
\newcommand{\tabincell}[2]{\begin{tabular}{@{}#1@{}}#2\end{tabular}}
\setlength{\abovecaptionskip}{4pt}
\setlength{\belowcaptionskip}{-10pt}
\footnotesize
\begin{tabular}{|c|c|}
\hline
 Loss Function & Decision Boundary \\
\hline\hline
Softmax Loss & $\thickmuskip=2mu \medmuskip=2mu (\bm{W}_{1} - \bm{W}_{2})\bm{x} + b_{1} - b_{2} = 0$\\\hline
Modified Softmax Loss & $\thickmuskip=2mu \medmuskip=2mu \|\bm{x}\|(\cos \theta_1 -\cos \theta_2) = 0$\\\hline
A-Softmax Loss & \tabincell{c}{$\thickmuskip=2mu \medmuskip=2mu \|\bm{x}\|(\cos m\theta_1 -\cos \theta_2) = 0$ for class 1\\$\thickmuskip=2mu \medmuskip=2mu \|\bm{x}\|(\cos \theta_1 -\cos m\theta_2) = 0$ for class 2}\\
\hline
\end{tabular}
\caption{\footnotesize Comparison of decision boundaries in binary case. Note that, $\theta_i$ is the angle between $\bm{W}_i$ and $\bm{x}$.}\label{decision}
\end{table}
\par
The justification of A-Softmax loss can also be made from decision boundary perspective. A-Softmax loss adopts different decision boundary for different class (each boundary is more stringent than the original), thus producing angular margin. The comparison of decision boundaries is given in Table~\ref{decision}. From original softmax loss to modified softmax loss, it is from optimizing inner product to optimizing angles. From modified softmax loss to A-Softmax loss, it makes the decision boundary more stringent and separated. The angular margin increases with larger $m$ and be zero if $\thickmuskip=2mu m=1$.
\par
Supervised by A-Softmax loss, CNNs learn face features with geometrically interpretable angular margin. Because A-Softmax loss requires $\thickmuskip=2mu \medmuskip=2mu \bm{W}_i=1,b_i=0$, it makes the prediction only depends on angles between the sample $\bm{x}$ and $\bm{W}_i$. So $\bm{x}$ can be classified to the identity with smallest angle. The parameter $m$ is added for the purpose of learning an angular margin between different identities.
\par
To facilitate gradient computation and back propagation, we replace $\cos(\theta_{j,i})$ and $\cos(m\theta_{y_i,i})$ with the expressions only containing $\bm{W}$ and $\bm{x}_i$, which is easily done by definition of cosine and multi-angle formula (also the reason why we need $m$ to be an integer). Without $\theta$, we can compute derivative with respect to $\bm{x}$ and $\bm{W}$, similar to softmax loss.
\subsection{Hypersphere Interpretation of A-Softmax Loss}
\label{geo}
\begin{figure}[t]
  \centering
  \renewcommand{\captionlabelfont}{\footnotesize}
  \setlength{\abovecaptionskip}{4pt}
  \setlength{\belowcaptionskip}{-7pt}
  \includegraphics[width=3.15in]{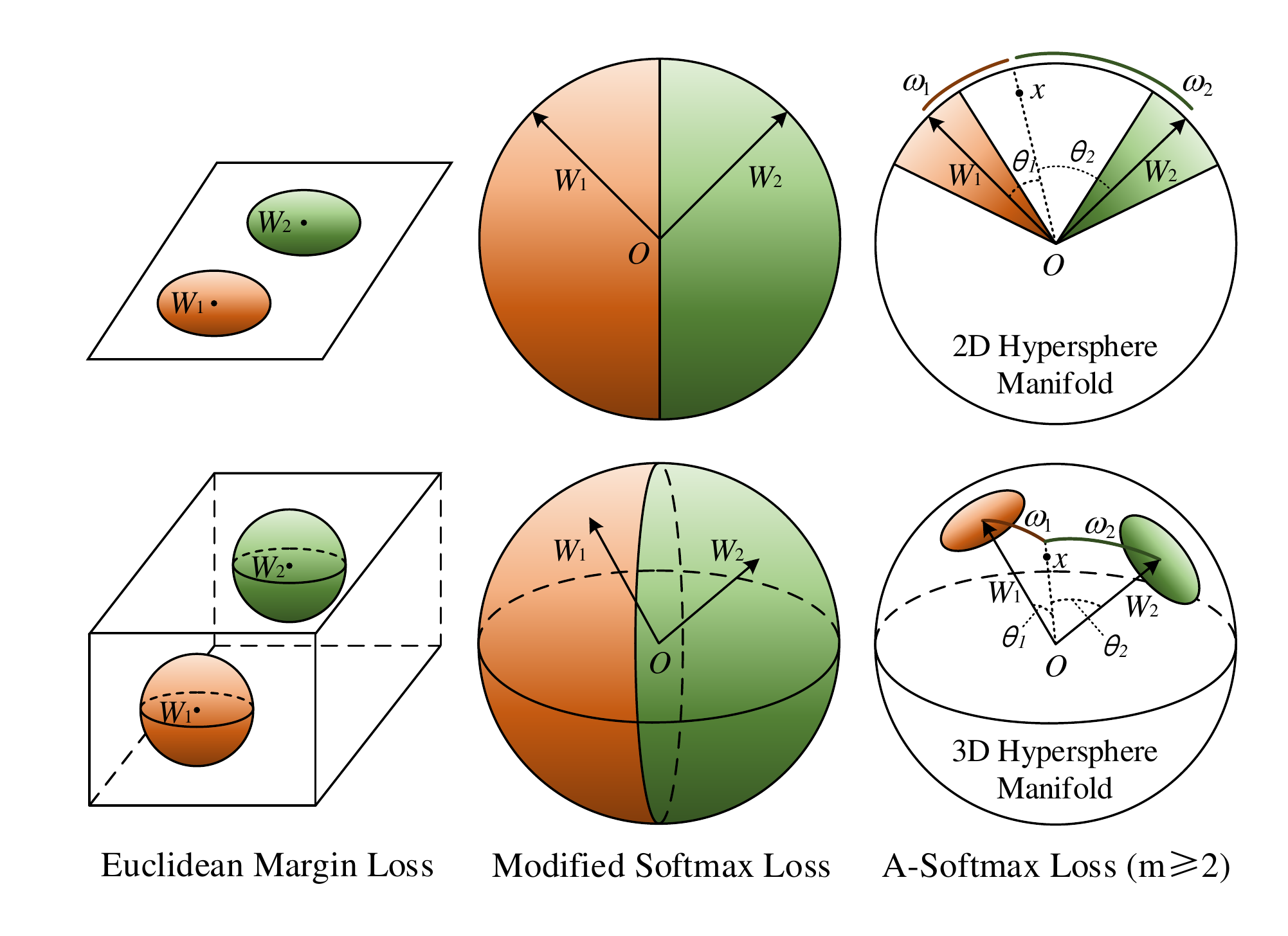}
  \caption{\footnotesize Geometry Interpretation of Euclidean margin loss (e.g. contrastive loss, triplet loss, center loss, etc.), modified softmax loss and A-Softmax loss. The first row is 2D feature constraint, and the second row is 3D feature constraint. The orange region indicates the discriminative constraint for class 1, while the green region is for class 2. \label{geoint}}
\end{figure}
A-Softmax loss has stronger requirements for a correct classification when $\thickmuskip=2mu m\geq2$, which generates an angular classification margin between learned features of different classes. A-Softmax loss not only imposes discriminative power to the learned features via angular margin, but also renders nice and novel hypersphere interpretation. As shown in Fig.~\ref{geoint}, A-Softmax loss is equivalent to learning features that are discriminative on a hypersphere manifold, while Euclidean margin losses learn features in Euclidean space.
\par
To simplify, We take the binary case to analyze the hypersphere interpretation. Considering a sample $\bm{x}$ from class 1 and two column weights $\bm{W}_1,\bm{W}_2$, the classification rule for A-Softmax loss is $\thickmuskip=2mu \medmuskip=2mu \cos (m\theta_1)>\cos(\theta_2)$, equivalently $\thickmuskip=2mu \medmuskip=2mu m\theta_1<\theta_2$. Notice that $\theta_1,\theta_2$ are equal to their corresponding arc length $\omega_1,\omega_2$\footnote{$\omega_i$ is the shortest arc length (geodesic distance) between $\bm{W}_i$ and the projected point of sample $\bm{x}$ on the unit hypersphere, while the corresponding $\theta_i$ is the angle between $\bm{W}_i$  and $\bm{x}$.} on unit hypersphere $\thickmuskip=0mu \medmuskip=0mu \{v_j,\forall j|\sum_j v_j^2 =1, v\geq 0\}$. Because $\thickmuskip=2mu \medmuskip=2mu \|\bm{W}\|_1=\|\bm{W}\|_2=1$, the decision replies on the arc length $\omega_1$ and $\omega_2$. The decision boundary is equivalent to $\thickmuskip=2mu \medmuskip=2mu m\omega_1=\omega_2$, and the constrained region for correctly classifying $\bm{x}$ to class 1 is $\thickmuskip=2mu \medmuskip=2mu m\omega_1<\omega_2$. Geometrically speaking, this is a hypercircle-like region lying on a hypersphere manifold. For example, it is a circle-like region on the unit sphere in 3D case, as illustrated in Fig.~\ref{geoint}. Note that larger $m$ leads to smaller hypercircle-like region for each class, which is an explicit discriminative constraint on a manifold. For better understanding, Fig.~\ref{geoint} provides 2D and 3D visualizations. One can see that A-Softmax loss imposes arc length constraint on a unit circle in 2D case and circle-like region constraint on a unit sphere in 3D case. Our analysis shows that optimizing angles with A-Softmax loss essentially makes the learned features more discriminative on a hypersphere.
\subsection{Properties of A-Softmax Loss}
\label{prop}
\begin{property}
A-Softmax loss defines a large angular margin learning task with adjustable difficulty. With larger $m$, the angular margin becomes larger, the constrained region on the manifold becomes smaller, and the corresponding learning task also becomes more difficult.
\end{property}
We know that the larger $m$ is, the larger angular margin A-Softmax loss constrains. There exists a minimal $m$ that constrains the maximal intra-class angular distance to be smaller than the minimal inter-class angular distance, which can also be observed in our experiments.
\begin{definition}[minimal $m$ for desired feature distribution]
$m_{\textnormal{min}}$ is the minimal value such that while $m>m_{\textnormal{min}}$, A-Softmax loss defines a learning task where the maximal intra-class angular feature  distance is constrained to be smaller than the minimal inter-class angular feature distance.
\end{definition}
\begin{property}[lower bound of $m_{\textnormal{min}}$ in binary-class case]
In binary-class case, we have $\thickmuskip=2mu \medmuskip=2mu m_{\textnormal{min}}\geq2+\sqrt{3}$.
\end{property}
\begin{proof}
We consider the space spaned by $\bm{W}_1$ and $\bm{W}_2$. Because $\thickmuskip=2mu m\geq2$, it is easy to obtain the maximal angle that class 1 spans is $\frac{\theta_{12}}{m-1}+\frac{\theta_{12}}{m+1}$ where $\theta_{12}$ is the angle between $\bm{W}_1$ and $\bm{W}_2$. To require the maximal intra-class feature angular distance smaller than the minimal inter-class feature angular distance, we need to constrain
\begin{equation}
\footnotesize
\underbrace{\frac{\theta_{12}}{m-1}+\frac{\theta_{12}}{m+1}}_{\text{max intra-class angle}}\leq\underbrace{\frac{(m-1)\theta_{12}}{m+1}}_{\text{min inter-class angle}},\ \  \theta_{12}\leq\frac{m-1}{m}\pi
\end{equation}
\begin{equation}
\footnotesize
\underbrace{\frac{2\pi-\theta_{12}}{m+1}+\frac{\theta_{12}}{m+1}}_{\text{max intra-class angle}}\leq\underbrace{\frac{(m-1)\theta_{12}}{m+1}}_{\text{min inter-class angle}},\ \  \theta_{12}>\frac{m-1}{m}\pi
\end{equation}
After solving these two inequalities, we could have $\thickmuskip=2mu \medmuskip=2mu m_{\textnormal{min}}\geq2+\sqrt{3}$, which is a lower bound for binary case.
\end{proof}
\par
\begin{property}[lower bound of $m_{\textnormal{min}}$ in multi-class case]
Under the assumption that $\bm{W}_i,\forall i$ are uniformly spaced in the Euclidean space, we have $m_{\textnormal{min}}\geq3$.
\end{property}
\begin{proof}
We consider the 2D $k$-class ($k\geq3$) scenario for the lower bound. Because $\bm{W}_i,\forall i$ are uniformly spaced in the 2D Euclidean space, we have $\thickmuskip=2mu \medmuskip=2mu \theta_{i}^{i+1}=\frac{2\pi}{k}$ where $\theta_{i}^{i+1}$ is the angle between $\bm{W}_i$ and $\bm{W}_{i+1}$. Since $\bm{W}_i,\forall i$ are symmetric, we only need to analyze one of them. For the $i$-th class ($\bm{W}_i$), We need to constrain
\begin{equation}
\footnotesize
\underbrace{\frac{\theta_{i}^{i+1}}{m+1}+\frac{\theta_{i-1}^{i}}{m+1}}_{\text{max intra-class angle}}\leq\underbrace{\min\bigg{\{}\frac{(m-1)\theta_{i}^{i+1}}{m+1},\frac{(m-1)\theta_{i-1}^{i}}{m+1}\bigg{\}}}_{\text{min inter-class angle}}
\end{equation}
After solving this inequality, we obtain $m_{\textnormal{min}}\geq3$, which is a lower bound for multi-class case.
\end{proof}
Based on this, we use $\thickmuskip=2mu \medmuskip=2mu m=4$ to approximate the desired feature distribution criteria. Since the lower bounds are not necessarily tight, giving a tighter lower bound and a upper bound under certain conditions is also possible, which we leave to the future work. Experiments also show that larger $m$ consistently works better and $\thickmuskip=2mu m=4$ will usually suffice.
\vspace{-.1mm}
\subsection{Discussions}
\vspace{-.1mm}
\textbf{Why angular margin.} First and most importantly, angular margin directly links to discriminativeness on a manifold, which intrinsically matches the prior that faces also lie on a manifold. Second, incorporating angular margin to softmax loss is actually a more natural choice. As Fig.~\ref{comp} shows, features learned by the original softmax loss have an intrinsic angular distribution. So directly combining Euclidean margin constraints with softmax loss is not reasonable.
\par
\textbf{Comparison with existing losses.} In deep FR task, the most popular and well-performing loss functions include contrastive loss, triplet loss and center loss. First, they only impose Euclidean margin to the learned features (w/o normalization), while ours instead directly considers angular margin which is naturally motivated. Second, both contrastive loss and triplet loss suffer from data expansion when constituting the pairs/triplets from the training set, while ours requires no sample mining and imposes discriminative constraints to the entire mini-batches (compared to contrastive and triplet loss that only affect a few representative pairs/triplets).
\par

\begin{table*}[t]
\renewcommand{\captionlabelfont}{\footnotesize}
\newcommand{\tabincell}[2]{\begin{tabular}{@{}#1@{}}#2\end{tabular}}
\centering
\setlength{\abovecaptionskip}{3pt}
\setlength{\belowcaptionskip}{-10pt}
\footnotesize
\begin{tabular}{|c|c|c|c|c|c|}
\hline
Layer & 4-layer CNN & 10-layer CNN & 20-layer CNN & 36-layer CNN & 64-layer CNN\\
\hline\hline
Conv1.x & [3$\times$3, 64]$\times$1, S2 & [3$\times$3, 64]$\times$1, S2 & \tabincell{c}{[3$\times$3, 64]$\times$1, S2\\$\left[\begin{aligned}&3\times3, 64\\&3\times3, 64\end{aligned}\right]\times 1$}& \tabincell{c}{[3$\times$3, 64]$\times$1, S2\\$\left[\begin{aligned}&3\times3, 64\\&3\times3, 64\end{aligned}\right]\times 2$} & \tabincell{c}{[3$\times$3, 64]$\times$1, S2\\$\left[\begin{aligned}&3\times3, 64\\&3\times3, 64\end{aligned}\right]\times 3$} \\\hline
Conv2.x  & [3$\times$3, 128]$\times$1, S2 & \tabincell{c}{[3$\times$3, 128]$\times$1, S2\\$\left[\begin{aligned}&3\times3, 128\\&3\times3, 128\end{aligned}\right]\times 1$} & \tabincell{c}{[3$\times$3, 128]$\times$1, S2\\$\left[\begin{aligned}&3\times3, 128\\&3\times3, 128\end{aligned}\right]\times 2$} & \tabincell{c}{[3$\times$3, 128]$\times$1, S2\\$\left[\begin{aligned}&3\times3, 128\\&3\times3, 128\end{aligned}\right]\times 4$} & \tabincell{c}{[3$\times$3, 128]$\times$1, S2\\$\left[\begin{aligned}&3\times3, 128\\&3\times3, 128\end{aligned}\right]\times 8$} \\\hline
Conv3.x  & [3$\times$3, 256]$\times$1, S2 & \tabincell{c}{[3$\times$3, 256]$\times$1, S2\\$\left[\begin{aligned}&3\times3, 256\\&3\times3, 256\end{aligned}\right]\times 2$} & \tabincell{c}{[3$\times$3, 256]$\times$1, S2\\$\left[\begin{aligned}&3\times3, 256\\&3\times3, 256\end{aligned}\right]\times 4$} & \tabincell{c}{[3$\times$3, 256]$\times$1, S2\\$\left[\begin{aligned}&3\times3, 256\\&3\times3, 256\end{aligned}\right]\times 8$} & \tabincell{c}{[3$\times$3, 256]$\times$1, S2\\$\left[\begin{aligned}&3\times3, 256\\&3\times3, 256\end{aligned}\right]\times 16$} \\\hline
Conv4.x  & [3$\times$3, 512]$\times$1, S2 & [3$\times$3, 512]$\times$1, S2 & \tabincell{c}{[3$\times$3, 512]$\times$1, S2\\$\left[\begin{aligned}&3\times3, 512\\&3\times3, 512\end{aligned}\right]\times 1$} & \tabincell{c}{[3$\times$3, 512]$\times$1, S2\\$\left[\begin{aligned}&3\times3, 512\\&3\times3, 512\end{aligned}\right]\times 2$} & \tabincell{c}{[3$\times$3, 512]$\times$1, S2\\$\left[\begin{aligned}&3\times3, 512\\&3\times3, 512\end{aligned}\right]\times 3$} \\\hline
FC1  & 512 & 512 & 512 & 512 & 512 \\\hline
\end{tabular}
\caption{\footnotesize Our CNN architectures with different convolutional layers. Conv1.x, Conv2.x and Conv3.x denote convolution units that may contain multiple convolution layers and residual units are shown in double-column brackets. E.g., [3$\times$3, 64]$\times$4 denotes 4 cascaded convolution layers with 64 filters of size 3$\times$3, and S2 denotes stride 2. FC1 is the fully connected layer. }\label{netarch}
\end{table*}
\vspace{-.75mm}
\section{Experiments (more in Appendix)}
\vspace{-.25mm}
\subsection{Experimental Settings}
\vspace{-.5mm}
\textbf{Preprocessing}. We only use standard preprocessing. The face landmarks in all images are detected by MTCNN \cite{zhang2016joint}. The cropped faces are obtained by similarity transformation. Each pixel ($[0,255]$) in RGB images is normalized by subtracting 127.5 and then being divided by 128.
\par
\begin{figure}[h]
  \centering
  \renewcommand{\captionlabelfont}{\footnotesize}
  \setlength{\abovecaptionskip}{3pt}
  \setlength{\belowcaptionskip}{-7pt}
  \includegraphics[width=2.9in]{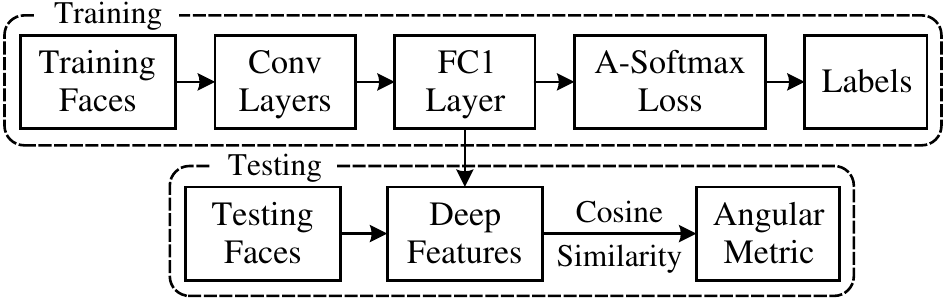}
  \caption{\footnotesize Training and Extracting \emph{SphereFace} features. \label{arch}}
\end{figure}

\textbf{CNNs Setup}. Caffe \cite{jia2014caffe} is used to implement A-Softmax loss and CNNs. The general framework to train and extract \emph{SphereFace} features is shown in Fig.~\ref{arch}. We use residual units \cite{he2016deep} in our CNN architecture. For fairness, all compared methods use the same CNN architecture (including residual units) as \emph{SphereFace}. CNNs with different depths (4, 10, 20, 36, 64) are used to better evaluate our method. The specific settings for difffernt CNNs we used are given in Table~\ref{netarch}. According to the analysis in Section \ref{prop}, we usually set $m$ as 4 in A-Softmax loss unless specified. These models are trained with batch size of 128 on four GPUs. The learning rate begins with 0.1 and is divided by 10 at the 16K, 24K iterations. The training is finished at 28K iterations.
\vspace{-1.9mm}
\par
\textbf{Training Data}. We use publicly available web-collected training dataset CASIA-WebFace \cite{yi2014learning} (after excluding the images of identities appearing in testing sets) to train our CNN models. CASIA-WebFace has 494,414 face images belonging to 10,575 different individuals. These face images are horizontally flipped for data augmentation. Notice that the scale of our training data (0.49M) is relatively small, especially compared to other private datasets used in DeepFace \cite{taigman2014deepface} (4M), VGGFace \cite{parkhi2015deep} (2M) and FaceNet \cite{schroff2015facenet} (200M).
\par
\textbf{Testing}. We extract the deep features (\emph{SphereFace}) from the output of the FC1 layer. For all experiments, the final representation of a testing face is obtained by concatenating its original face features and its horizontally flipped features. The score (metric) is computed by the cosine distance of two features. The nearest neighbor classifier and thresholding are used for face identification and verification, respectively.
\subsection{Exploratory Experiments}
\begin{figure*}[t]
  \centering
  \renewcommand{\captionlabelfont}{\footnotesize}
  \setlength{\abovecaptionskip}{3pt}
  \setlength{\belowcaptionskip}{-8pt}
  \includegraphics[width=6.43in]{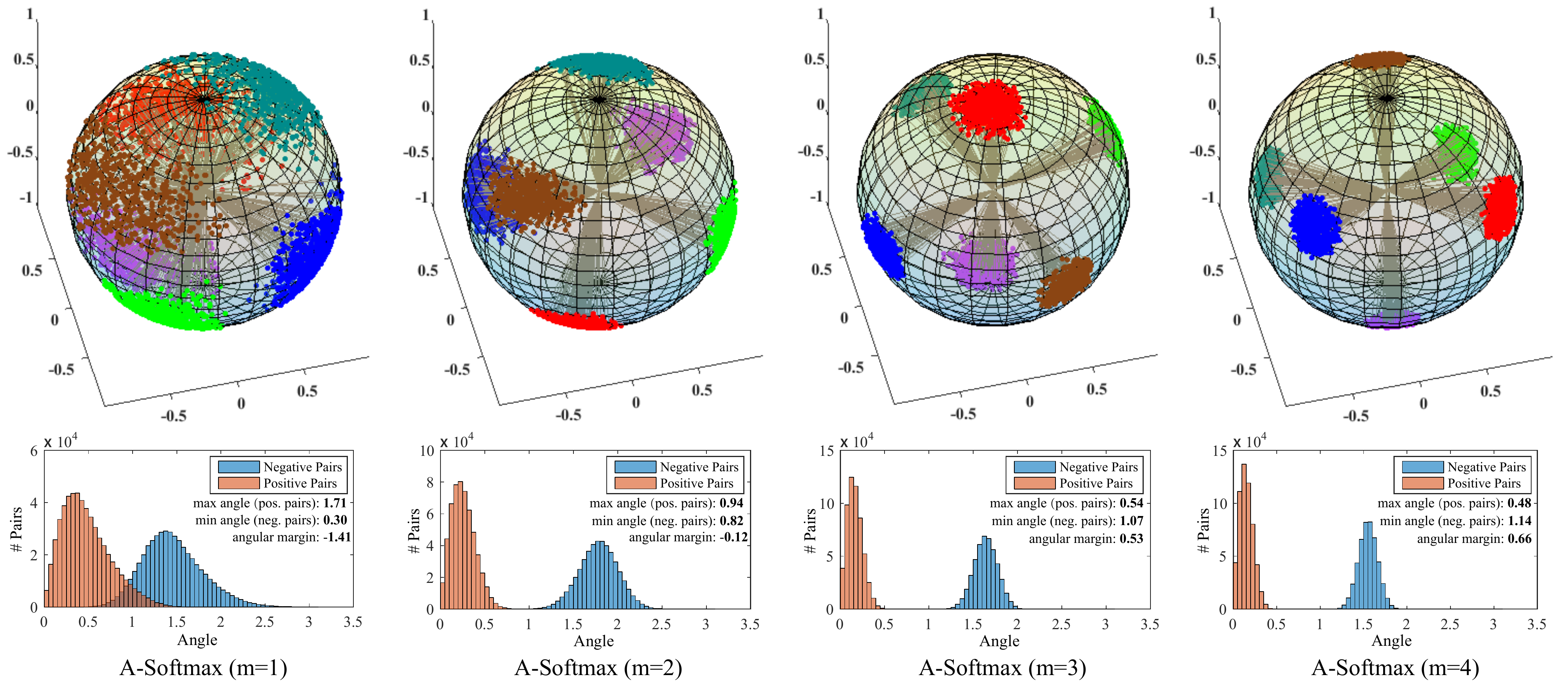}\\
  \caption{\footnotesize Visualization of features learned with different $m$. The first row shows the 3D features projected on the unit sphere. The projected points are the intersection points of the feature vectors and the unit sphere. The second row shows the angle distribution of both positive pairs and negative pairs (we choose class 1 and class 2 from the subset to construct positive and negative pairs). Orange area indicates positive pairs while blue indicates negative pairs. All angles are represented in radian. Note that, this visualization experiment uses a 6-class subset of the CASIA-WebFace dataset.}\label{vis_m}
\end{figure*}
\textbf{Effect of $\bm{m}$.} To show that larger $m$ leads to larger angular margin (i.e. more discriminative feature distribution on manifold), we perform a toy example with different $m$. We train A-Softmax loss with 6 individuals that have the most samples in CASIA-WebFace. We set the output feature dimension (FC1) as 3 and visualize the training samples in Fig.~\ref{vis_m}. One can observe that larger $m$ leads to more discriminative distribution on the sphere and also larger angular margin, as expected. We also use class 1 (blue) and class 2 (dark green) to construct positive and negative pairs to evaluate the angle distribution of features from the same class and different classes. The angle distribution of positive and negative pairs (the second row of Fig.~\ref{vis_m}) quantitatively shows the angular margin becomes larger while $m$ increases and every class also becomes more distinct with each other.
\par
Besides visual comparison, we also perform face recognition on LFW and YTF to evaluate the effect of $m$. For fair comparison, we use 64-layer CNN (Table~\ref{netarch}) for all losses. Results are given in Table~\ref{diffm}. One can observe that while $m$ becomes larger, the accuracy of A-Softmax loss also becomes better, which shows that  larger angular margin can bring stronger discrimination power.
\par
\vspace{-1.0mm}
\begin{table}[h]
\centering
\footnotesize
\renewcommand{\captionlabelfont}{\footnotesize}
\setlength{\abovecaptionskip}{3pt}
\setlength{\belowcaptionskip}{-7pt}
\begin{tabular}{|c|c|c|c|c|c|}
\hline
Dataset & Original & m=1 & m=2 & m=3 & m=4 \\
\hline\hline
LFW & 97.88 & 97.90 & 98.40 & 99.25 & \textbf{99.42} \\
YTF & 93.1 & 93.2 & 93.8 & 94.4 & \textbf{95.0} \\
\hline
\end{tabular}
\caption{\footnotesize Accuracy(\%) comparison of different $m$ (A-Softmax loss) and original softmax loss on LFW and YTF dataset.}\label{diffm}
\end{table}
\par
\begin{figure}[h]
  \centering
  \renewcommand{\captionlabelfont}{\footnotesize}
  \setlength{\abovecaptionskip}{3pt}
  \setlength{\belowcaptionskip}{-6pt}
  \includegraphics[width=3.2in]{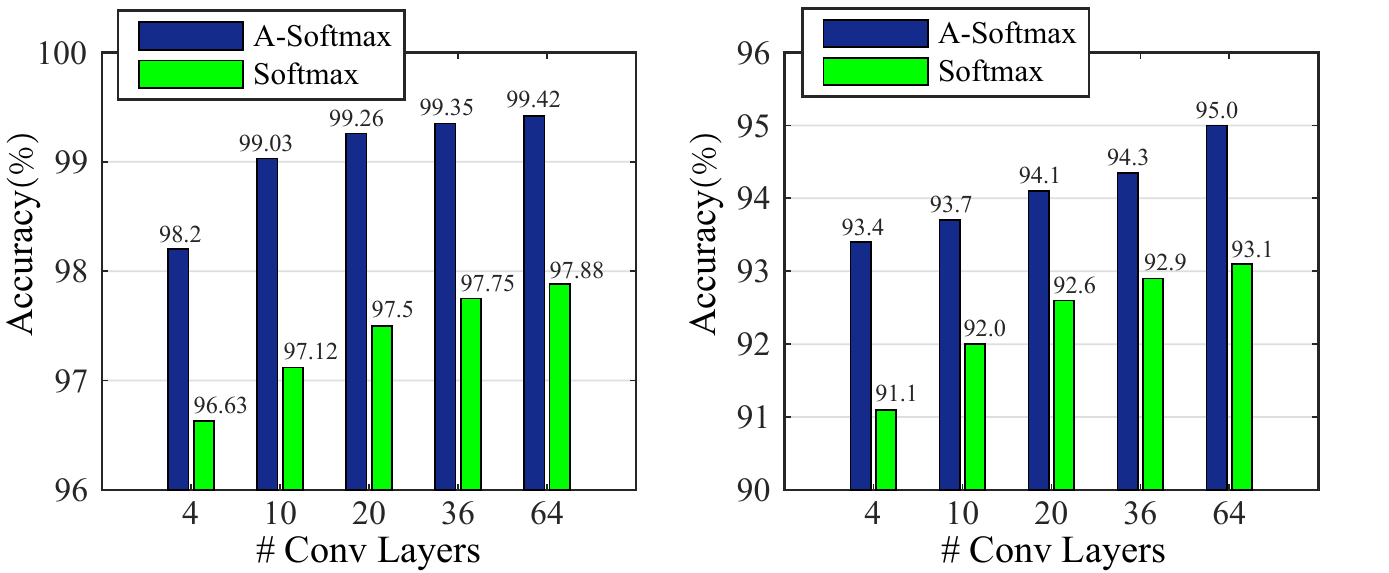}\\
  \caption{\footnotesize Accuracy (\%) on LFW and YTF with different number of convolutional layers. Left side is for LFW, while right side is for YTF.}\label{bar}
\end{figure}

\textbf{Effect of CNN architectures.} We train A-Softmax loss ($\thickmuskip=2mu m=4$) and original softmax loss with different number of convolution layers. Specific CNN architectures can be found in Table~\ref{netarch}. From Fig.~\ref{bar}, one can observe that A-Softmax loss consistently outperforms CNNs with softmax loss (1.54\%$\sim$1.91\%), indicating that A-Softmax loss is more suitable for open-set FR. Besides, the difficult learning task defined by A-Softmax loss makes full use of the superior learning capability of deeper architectures. A-Softmax loss greatly improve the verification accuracy from 98.20\% to 99.42\% on LFW, and from 93.4\% to 95.0\% on YTF. On the contrary, the improvement of deeper standard CNNs is unsatisfactory and also easily get saturated (from 96.60\% to 97.75\% on LFW, from 91.1\% to 93.1\% on YTF).
\par
\vspace{-1.2mm}
\subsection{Experiments on LFW and YTF}
\begin{table}[t]
\centering
\footnotesize
\renewcommand{\captionlabelfont}{\footnotesize}
\setlength{\abovecaptionskip}{6pt}
\setlength{\belowcaptionskip}{-10pt}
\begin{tabular}{|c|c|c|c|c|}
\hline
Method & Models & Data & LFW & YTF \\
\hline\hline
DeepFace \cite{taigman2014deepface} & 3 & 4M* & 97.35 & 91.4 \\
FaceNet \cite{schroff2015facenet} & 1 & 200M* & \textbf{99.65} & 95.1 \\
Deep FR \cite{parkhi2015deep} & 1 & 2.6M & 98.95 & \textbf{97.3}\\
DeepID2+ \cite{sun2015deeply} & 1 & 300K* & 98.70 & N/A \\
DeepID2+ \cite{sun2015deeply} & 25 & 300K* & 99.47 & 93.2 \\
Baidu \cite{liu2015targeting} & 1 & 1.3M* & 99.13 & N/A \\
Center Face \cite{wen2016discriminative} & 1 & 0.7M* & 99.28 & 94.9 \\\hline\hline
Yi et al. \cite{yi2014learning}& 1 & WebFace & 97.73 & 92.2 \\
Ding et al. \cite{ding2015robust}& 1 & WebFace & 98.43 & N/A \\
Liu et al. \cite{liu2016large}& 1 & WebFace & 98.71 & N/A \\\hline\hline
Softmax Loss & 1 & WebFace & 97.88 & 93.1\\
Softmax+Contrastive \cite{sun2014deep} & 1 & WebFace & 98.78 & 93.5\\
Triplet Loss \cite{schroff2015facenet} & 1 & WebFace & 98.70 & 93.4\\
L-Softmax Loss \cite{liu2016large}& 1 & WebFace & 99.10 & 94.0\\
Softmax+Center Loss \cite{wen2016discriminative} & 1 & WebFace & 99.05 & 94.4\\\hline\hline
SphereFace & 1 & WebFace & \textbf{99.42} & \textbf{95.0}\\
\hline
\end{tabular}
\caption{\footnotesize Accuracy (\%) on LFW and YTF dataset. * denotes the outside data is private (not publicly available). For fair comparison, all loss functions (including ours) we implemented use 64-layer CNN architecture in Table~\ref{netarch}.}\label{lfwytf}
\end{table}
LFW dataset \cite{huang2007labeled} includes 13,233 face images from 5749 different identities, and YTF dataset \cite{wolf2011face} includes 3,424 videos from 1,595 different individuals. Both datasets contains faces with large variations in pose, expression and illuminations. We follow the unrestricted with labeled outside data protocol \cite{huang2014labeled} on both datasets. The performance of \emph{SphereFace} are evaluated on 6,000 face pairs from LFW and 5,000 video pairs from YTF. The results are given in Table~\ref{lfwytf}. For contrastive loss and center loss, we follow the FR convention to form a weighted combination with softmax loss. The weights are selected via cross validation on training set. For L-Softmax \cite{liu2016large}, we also use $\thickmuskip=2mu m=4$. All the compared loss functions share the same 64-layer CNN architecture.
\par
\begin{figure*}[t]
  \centering
  \renewcommand{\captionlabelfont}{\footnotesize}
  \setlength{\abovecaptionskip}{3pt}
  \setlength{\belowcaptionskip}{-10pt}
  \includegraphics[width=6.6in]{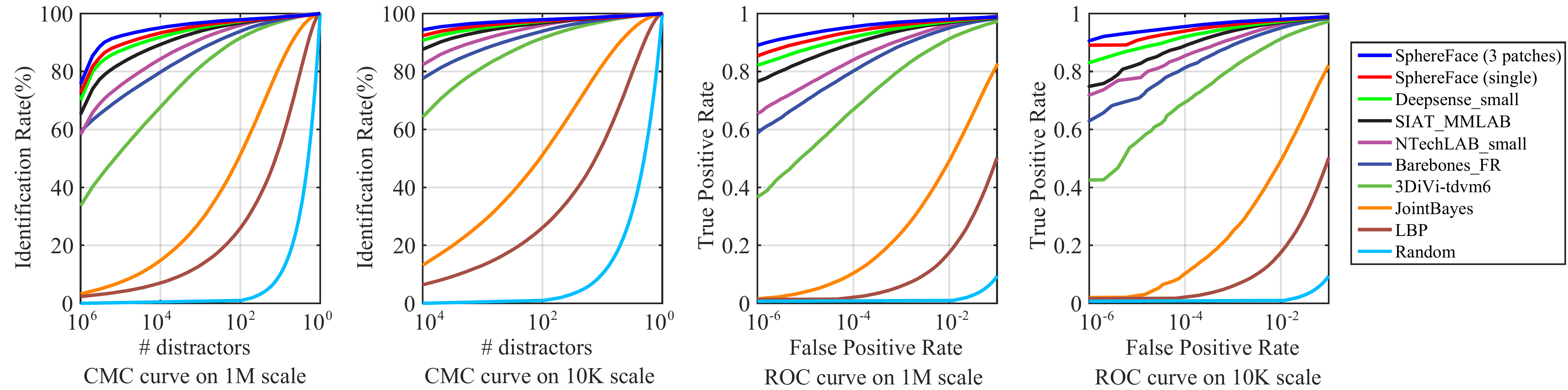}\\
  \caption{\footnotesize  CMC and ROC curves of different methods under the small training set protocol.}\label{megafacefig}
\end{figure*}
\par
Most of the existing face verification systems achieve high performance with huge training data or model ensemble. While using single model trained on publicly available dataset (CAISA-WebFace, relatively small and having noisy labels), \emph{SphereFace} achieves 99.42\% and 95.0\% accuracies on LFW and YTF datasets. It is the current best performance trained on WebFace and considerably better than the other models trained on the same dataset. Compared with models trained on high-quality private datasets, \emph{SphereFace} is still very competitive, outperforming most of the existing results in Table~\ref{lfwytf}. One should notice that our single model performance is only worse than Google FaceNet which is trained with more than 200 million data.
\par
For fair comparison, we also implement the softmax loss, contrastive loss, center loss, triplet loss, L-Softmax loss \cite{liu2016large} and train them with the same 64-layer CNN architecture as A-Softmax loss. As can be observed in Table~\ref{lfwytf}, \emph{SphereFace} consistently outperforms the features learned by all these compared losses, showing its superiority in FR tasks.
\subsection{Experiments on MegaFace Challenge}
\begin{table}[t]
\centering
\footnotesize
\renewcommand{\captionlabelfont}{\footnotesize}
\newcommand{\tabincell}[2]{\begin{tabular}{@{}#1@{}}#2\end{tabular}}
\setlength{\abovecaptionskip}{4pt}
\setlength{\belowcaptionskip}{-11pt}
\begin{tabular}{|c|c|c|c|c|}
\hline
Method & protocol & Rank1 Acc. & Ver. \\
\hline\hline
NTechLAB - facenx large & Large & 73.300 & 85.081 \\
Vocord - DeepVo1 & Large & \textbf{75.127} & 67.318 \\
Deepsense - Large & Large & 74.799 & \textbf{87.764} \\
Shanghai Tech & Large & 74.049 & 86.369 \\
Google - FaceNet v8 & Large & 70.496 & 86.473 \\
Beijing FaceAll\_Norm\_1600 & Large & 64.804 & 67.118\\
Beijing FaceAll\_1600 & Large & 63.977 & 63.960 \\\hline\hline
Deepsense - Small & Small & \textbf{70.983} & \textbf{82.851}\\
SIAT\_MMLAB & Small & 65.233 & 76.720 \\
Barebones FR - cnn & Small & 59.363 & 59.036 \\
NTechLAB - facenx\_small & Small & 58.218 & 66.366 \\
3DiVi Company - tdvm6 & Small & 33.705 & 36.927 \\\hline\hline
Softmax Loss & Small & 54.855 & 65.925 \\
Softmax+Contrastive Loss \cite{sun2014deep} & Small & 65.219 & 78.865 \\
Triplet Loss \cite{schroff2015facenet} & Small & 64.797 & 78.322 \\
L-Softmax Loss \cite{liu2016large} & Small & 67.128 & 80.423 \\
Softmax+Center Loss \cite{wen2016discriminative} & Small & 65.494 & 80.146 \\\hline\hline
SphereFace (single model) & Small & \textbf{72.729} & \textbf{85.561}\\
SphereFace (3-patch ensemble) & Small & \textbf{75.766} & \textbf{89.142}\\
\hline
\end{tabular}
\caption{\footnotesize Performance (\%) on MegaFace challenge. ``Rank-1 Acc.'' indicates rank-1 identification accuracy with 1M distractors, and ``Ver.'' indicates verification TAR for $10^{-6}$ FAR. TAR and FAR denote True Accept Rate and False Accept Rate respectively. For fair comparison, all loss functions (including ours) we implemented use the same deep CNN architecture.}\label{megaface}
\end{table}
MegaFace dataset \cite{miller2015megaface} is a recently released testing benchmark with very challenging task to evaluate the performance of face recognition methods at the million scale of distractors. MegaFace dataset contains a gallery set and a probe set. The gallery set contains more than 1 million images from 690K different individuals. The probe set consists of two existing datasets: Facescrub \cite{ng2014data} and FGNet. MegaFace has several testing scenarios including identification, verification and pose invariance under two protocols (large or small training set). The training set is viewed as small if it is less than 0.5M. We evaluate \emph{SphereFace} under the small training set protocol. We adopt two testing protocols: face identification and verification. The results are given in Fig.~\ref{megafacefig} and Tabel \ref{megaface}. Note that we use simple 3-patch feature concatenation ensemble as the final performance of \emph{SphereFace}.
\par
Fig.~\ref{megafacefig} and Tabel \ref{megaface} show that \emph{SphereFace} (3 patches ensemble) beats the second best result by a large margins (4.8\% for rank-1 identification rate and 6.3\% for verification rate) on MegaFace benchmark under the small training dataset protocol. Compared to the models trained on large dataset (500 million for Google and 18 million for NTechLAB), our method still performs better (0.64\% for id. rate and 1.4\% for veri. rate). Moreover, in contrast to their sophisticated network design, we only employ typical CNN architecture supervised by A-Softamx to achieve such excellent performance. For single model \emph{SphereFace}, the accuracy of face identification and verification are still 72.73\% and 85.56\% respectively, which already outperforms most state-of-the-art methods. For better evaluation, we also implement the softmax loss, contrastive loss, center loss, triplet loss and L-Softmax loss \cite{liu2016large}. Compared to these loss functions trained with the same CNN architecture and dataset, \emph{SphereFace} also shows significant and consistent improvements. These results convincingly demonstrate that the proposed \emph{SphereFace} is well designed for open-set face recognition. One can also see that learning features with large inter-class angular margin can significantly improve the open-set FR performance.
\vspace{-7mm}
\section{Concluding Remarks}
\vspace{-1.5mm}
This paper presents a novel deep hypersphere embedding approach for face recognition. In specific, we propose the angular softmax loss for CNNs to learn discriminative face features (\emph{SphereFace}) with angular margin. A-Softmax loss renders nice geometric interpretation by constraining learned features to be discriminative on a hypersphere manifold, which intrinsically matches the prior that faces also lie on a non-linear manifold. This connection makes A-Softmax very effective for learning face representation. Competitive results on several popular face benchmarks demonstrate the superiority and great potentials of our approach. We also believe A-Softmax loss could also benefit some other tasks like object recognition, person re-identification, etc.
\par

{
\small
\bibliographystyle{ieee}
\bibliography{example_paper}
}

\clearpage
\newpage

\appendix
\onecolumn

\begin{appendix}
	
	\begin{center}
		{\Large \bf Appendix}
	\end{center}

\section{The intuition of removing the last ReLU}
Standard CNNs usually connect ReLU to the bottom of FC1, so the learned features will only distribute in the non-negative range $[0,+\infty)$, which limits the feasible learning space (angle) for the CNNs. To address this shortcoming, both SphereFace and \cite{liu2016large} first propose to remove the ReLU nonlinearity that is connected to the bottom of FC1 in SphereFace networks. Intuitively, removing the ReLU can greatly benefit the feature learning, since it provides larger feasible learning space (from angular perspective).
\par
\begin{figure*}[h]
	\centering
	\renewcommand{\captionlabelfont}{\footnotesize}
	\setlength{\abovecaptionskip}{2pt}
	\setlength{\belowcaptionskip}{-4pt}
	\includegraphics[width=6.2in]{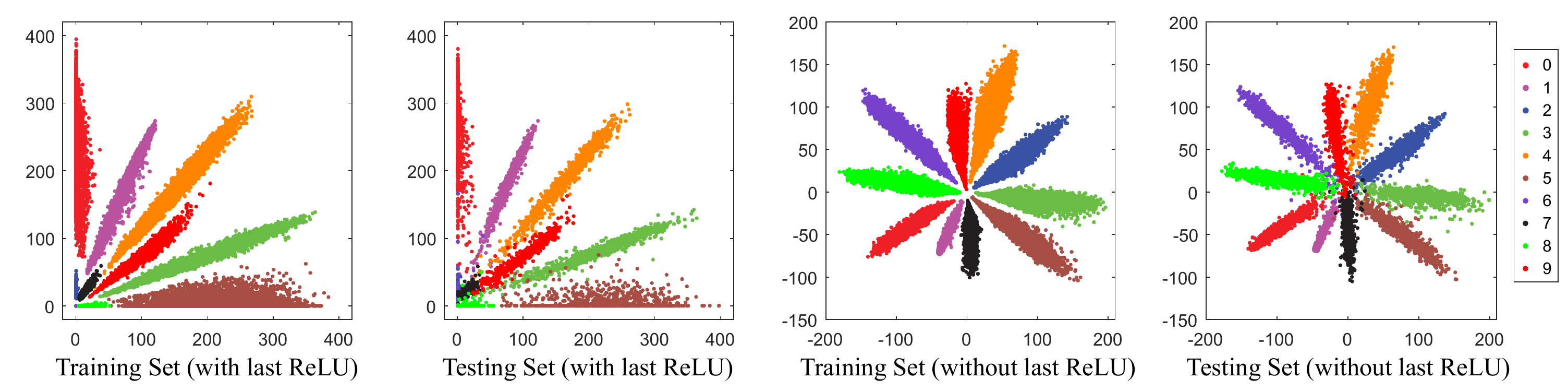}\\
	\caption{\footnotesize 2-D visualization before and after removing the last ReLU.}\label{nonlinear}
\end{figure*}

\textbf{Visualization on MNIST}. Fig. \ref{nonlinear} shows the 2-D visualization of feature distributions in MNIST with and without the last ReLU. One can observe with ReLU the 2-D feature could only distribute in the first quadrant. Without the last ReLU, the learned feature distribution is much more reasonable.
\par
\section{Normalizing the weights could reduce the prior caused by the training data imbalance}
We have emphasized in the main paper that normalizing the weights can give better geometric interpretation. Besides this, we also justify why we want to normalize the weights from a different perspective. We find that normalizing the weights can implicitly reduce the prior brought by the training data imbalance issue (e.g., the long-tail distribution of the training data). In other words, we argue that normalizing the weights can partially address the training data imbalance problem.
\par
\begin{figure*}[h]
	\centering
	\renewcommand{\captionlabelfont}{\footnotesize}
	\setlength{\abovecaptionskip}{2pt}
	\setlength{\belowcaptionskip}{5pt}
	\includegraphics[width=6.2in]{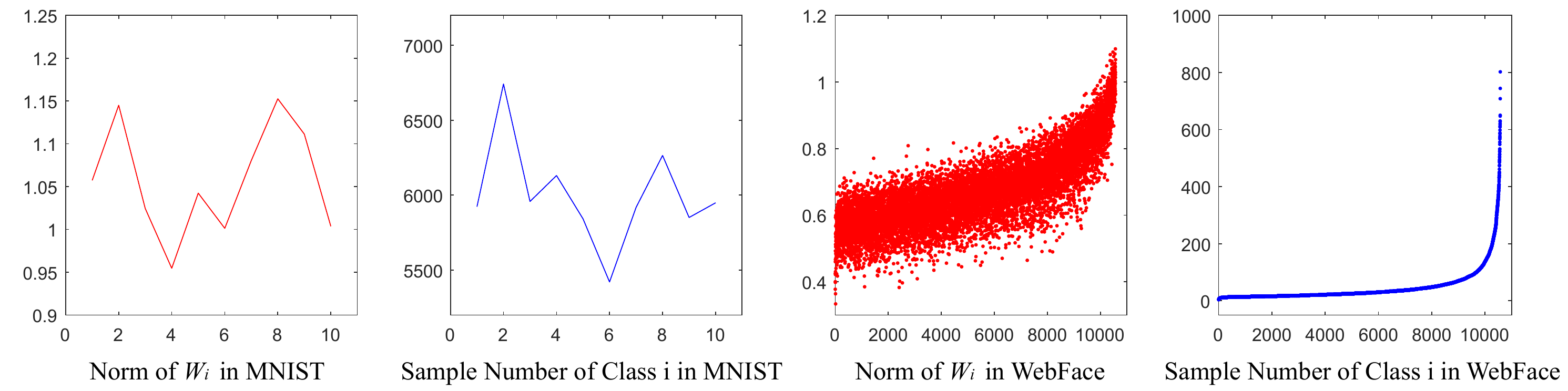}
	\caption{\footnotesize Norm of $\bm{W}_i$ and sample number of class $i$ in MNIST dataset and  CASIA-WebFace dataset.}\label{norm1}
\end{figure*}
We have an empirical study on the relation between the sample number of each class and the 2-norm of the weights corresponding to the same class (the $i$-th column of $\bm{W}$ is associated to the $i$-th class). By computing the norm of $\bm{W}_i$ and sample number of class $i$ with respect to each class (see Fig. \ref{norm1}), we find that the larger sample number a class has, the larger the associated norm of weights tends to be. We argue that the norm of weights $\bm{W}_i$ with respect to class $i$ is largely determined by its sample distribution and sample number. Therefore, norm of weights $\bm{W}_i, \forall i$ can be viewed as a learned prior hidden in training datasets. Eliminating such prior is often beneficial to face verification. This is because face verification requires to test on a dataset whose idenities can not appear in training datasets, so the prior from training dataset should not be transferred to the testing. This prior may even be harmful to face verification performance. To eliminate such prior, we normalize the norm of weights of FC2\footnote{FC2 refers to the fully connected layer in the softmax loss (or A-Softmax loss).}.
\section{Empirical experiment of zeroing out the biases}
\begin{figure*}[h]
	\centering
	\renewcommand{\captionlabelfont}{\footnotesize}
	\setlength{\abovecaptionskip}{0pt}
	\setlength{\belowcaptionskip}{-5pt}
	\includegraphics[width=3.8in]{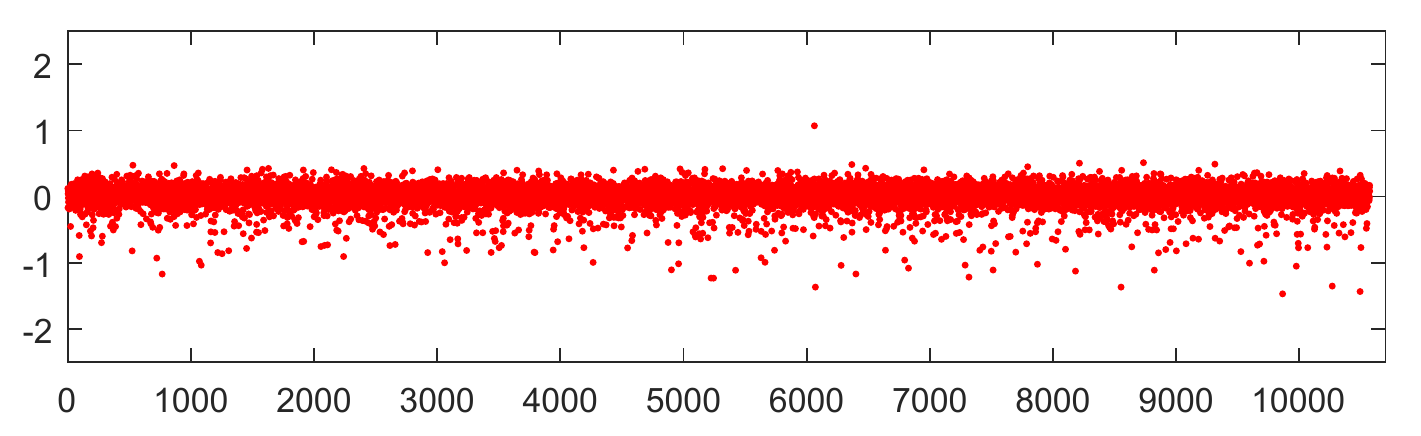}
	\caption{\footnotesize Biases of last fully connected layer learned in CASIA-WebFace dataset.}\label{bias2}
\end{figure*}
Standard CNNs usually preserve the bias term in the fully connected layers, but these bias terms make it difficult to analyze the proposed A-Softmax loss. This is because SphereFace aims to optimize the angle and produce the angular margin. With bias of FC2, the angular geometry interpretation becomes much more difficult to analyze. To facilitate the analysis, we zero out the bias of FC2 following \cite{liu2016large}. By setting the bias of FC2 to zero, the A-Softmax loss has clear geometry interpretation and therefore becomes much easier to analyze. We show all the biases of FC2 from a CASIA-pretrained model in Fig. \ref{bias2}. One can observe that the most of the biases are near zero, indicating these biases are not necessarily useful for face verification.
\par
\begin{figure*}[h]
	  \centering
	  \renewcommand{\captionlabelfont}{\footnotesize}
	  \setlength{\abovecaptionskip}{2pt}
	  \setlength{\belowcaptionskip}{-5pt}
	  \includegraphics[width=6.5in]{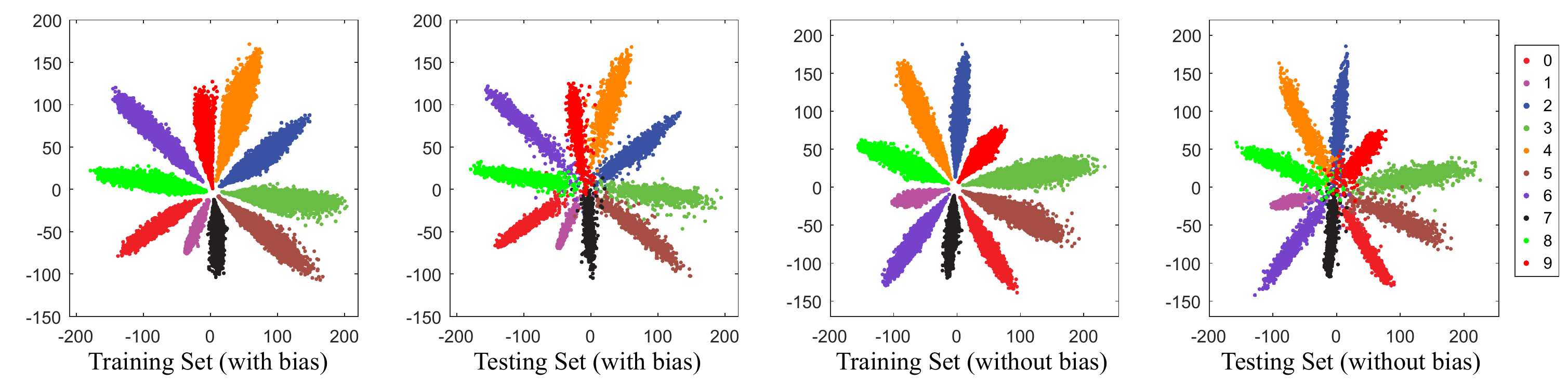}
	  \caption{\footnotesize 2-D visualization with and without bias of last fully connected layer in MNIST.}\label{bias}
\end{figure*}
\textbf{Visualization on MNIST}. We visualize the 2-D feature distribution in MNIST dataset with and without bias in Fig. \ref{bias}. One can observe that zeroing out the bias has no direct influence on the feature distribution. The features learned with and without bias can both make full use of the learning space.

\section{2D visualization of A-Softmax loss on MNIST}
\begin{figure}[h]
	\centering
	\renewcommand{\captionlabelfont}{\footnotesize}
	\setlength{\abovecaptionskip}{2pt}
	\setlength{\belowcaptionskip}{-3pt}
	\includegraphics[width=6.5in]{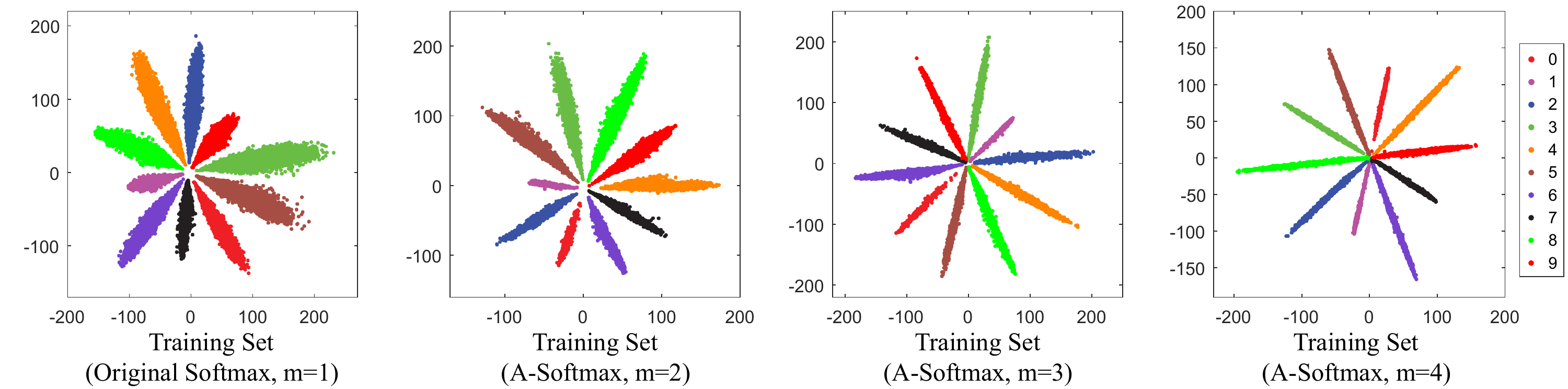}\\
	\caption{\footnotesize 2-D MNIST visualization of features learned by the softmax loss and the A-Softmax loss ($m=2,3,4$).}\label{loss}
\end{figure}

We visualize the 2-D feature distribution on MNIST in Fig. \ref{loss}. It is obvious that with larger $m$ the learned features become much more discriminative due to the larger inter-class angular margin. Most importantly, the learned discriminative features also generalize really well in the testing set.
\par
\section{Angular Fisher score for evaluating the feature discriminativeness and ablation study on our proposed modifications}
We first propose an angular Fisher score for evaluating the feature discriminativeness in angular margin feature learning. The angular Fisher score (AFS) is defined by

\begin{equation}
AFS=\frac{S_w}{S_b}
\end{equation}
where the within-class scatter value is defined as $S_w=\sum_i\sum_{x_j\in X_i}(1-\cos\langle x_j,m_i\rangle)$ and the between-class scatter value is defined as $S_b=\sum_in_i(1-\cos\langle m_i,m\rangle)$. $X_i$ is the $i$-th class samples, $m_i$ is the mean vector of features from class $i$, $m$ is the mean vector of the whole dataset, and $n_i$ is the sample number of class $i$. In general, the lower the fisher value is, the more discriminative the features are.
\par
Next, we perform a comprehensive ablation study on all the proposed modifications: removing last ReLU, removing Biases, normalizing weights and applying A-Softmax loss. The experiments are performed using the 4-layer CNN described in Table \ref{netarch}. The models are trained on CASIA dataset and tested on LFW dataset. The setting is exactly the same as the LFW experiment in the main paper. As shown in Table \ref{ablation_LFW}, we could observe that all our modification leads to peformance improvement and our A-Softmax could greatly increase the angular feature discriminativeness.
\begin{table}[h]
	\centering
	\renewcommand{\captionlabelfont}{\footnotesize}
	\setlength{\abovecaptionskip}{5pt}
	\setlength{\belowcaptionskip}{-2pt}
	\footnotesize
	\begin{tabular}{|c|c|c|c|c|c|c|}
		\hline
		CNN & Remove Last ReLU & Remove Biases & Normalize Weights & A-Softmax & Accuracy & Angular Fisher Score \\
		\hline\hline
		A & No & No & No & No & 95.13 & 0.3477\\
		B & Yes & No & No & No & 96.37 & 0.2835\\
		C & Yes & Yes & No & No & 96.40 & 0.2815 \\
		D & Yes & Yes & Yes & No & 96.63 & 0.2462 \\
		E & Yes & Yes & Yes & Yes (m=2) & 97.67 & 0.2277\\
		F & Yes & Yes & Yes & Yes (m=3) & 97.82 & 0.1791\\
		G & Yes & Yes & Yes & Yes (m=4) & \textbf{98.20} & \textbf{0.1709} \\
		\hline
	\end{tabular}
	\caption{\footnotesize Verification accuracy (\%) on LFW dataset.}\label{ablation_LFW}
\end{table}
\section{Experiments on MegaFace with different convolutional layers}
We also perform the experiment on MegaFace dataset with CNN of different convolutional layers. The results in Table \ref{megaface_appendix} show that the A-Softmax loss could make best use of the network capacity. With more convolutional layers, the A-Softmax loss (i.e., SphereFace) performs better. Most notably, SphereFace with only 4 convolutional layer could peform better than the softmax loss with 64 convolutional layers, which validates the superiority of our A-Softmax loss.
\begin{table}[h]
	\centering
	\renewcommand{\captionlabelfont}{\footnotesize}
	\footnotesize
	\newcommand{\tabincell}[2]{\begin{tabular}{@{}#1@{}}#2\end{tabular}}
	\setlength{\abovecaptionskip}{5pt}
	\setlength{\belowcaptionskip}{-3pt}
	\begin{tabular}{|c|c|c|c|c|}
		\hline
		Method & protocol & \tabincell{c}{Rank-1 Id. Acc.\\with 1M distractors} & \tabincell{c}{Ver. TAR\\for $10^{-6}$ FAR} \\
		\hline\hline
		Softmax Loss (64 conv layers) & Small & 54.855 & 65.925\\
		SphereFace (4 conv layers) & Small & 57.529 & 68.547\\
		SphereFace (10 conv layers) & Small & 65.335 & 78.069\\
		SphereFace (20 conv layers) & Small & 69.623 & 83.159\\
		SphereFace (36 conv layers) & Small & 71.257 & 84.052\\
		SphereFace (64 conv layers) & Small & \textbf{72.729} & \textbf{85.561}\\
		\hline
	\end{tabular}
	\caption{\footnotesize Performance (\%) on MegaFace challenge with different convolutional layers. TAR and FAR denote True Accept Rate and False Accept Rate respectively. For all the SphereFace models, we use $m=4$. With larger $m$ and proper network optimization, the performance could potentially keep increasing.}\label{megaface_appendix}
\end{table}
\section{The annealing optimization strategy for A-Softmax loss}
The optimization of the A-Softmax loss is similar to the L-Softmax loss~\cite{liu2016large}. We use an annealing optimization strategy to train the network with A-Softmax loss. To be simple, the annealing strategy is essentially supervising the newtork from an easy task (i.e., large $\lambda$) gradually to a difficult task (i.e., small $\lambda$). Specifically, we let  $f_{y_i}=\frac{\lambda\|\bm{x}_i\|\cos(\theta_{y_i})+\|\bm{x}_i\|\psi({\theta_{y_i}})}{1+\lambda}$ and start the stochastic gradient descent initially with a very large $\lambda$ (it is equivalent to optimizing the original softmax). Then we gradually reduce $\lambda$ during training. Ideally $\lambda$ can be gradually reduced to zero, but in practice, a small value will usually suffice. In most of our face experiments, decaying $\lambda$ to 5 has already lead to impressive results. Smaller $\lambda$ could potentially yield a better performance but is also more difficult to train.
\section{Details of the 3-patch ensemble strategy in MegaFace challenge}
\begin{figure}[h]
	\centering
	\renewcommand{\captionlabelfont}{\footnotesize}
	\setlength{\abovecaptionskip}{2pt}
	\setlength{\belowcaptionskip}{-3pt}
	\includegraphics[width=4in]{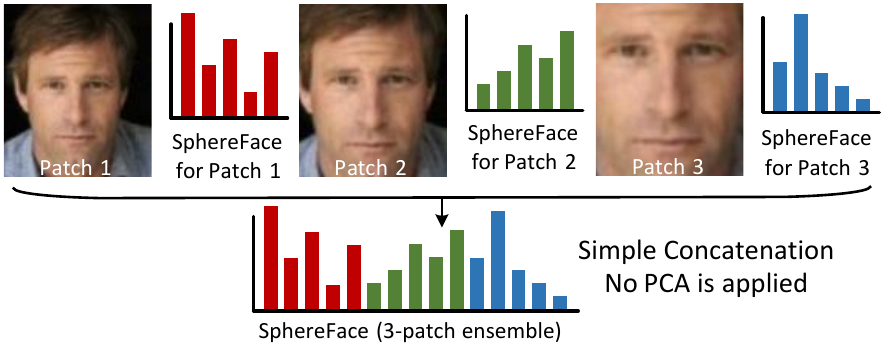}\\
	\caption{\footnotesize 3-Patch ensembles in SphereFace for MegaFace challenge.}\label{ensemble}
\end{figure}

We adopt a common strategy to perform the 3-patch ensemble, as shown in Fig.~\ref{ensemble}. Although using more patches could keep increasing the performance, but considering the tradeoff between efficiency and accuracy, we use 3-patch simple concatenation ensemble (without the use of PCA). The 3 patches can be selected by cross-validation. The 3 patches we use in the paper are exactly the same as in Fig.~\ref{ensemble}.
\end{appendix}

\end{document}